\pdfoutput=1

\documentclass[11pt]{article}

\usepackage[preprint]{acl}
\usepackage{multirow}
\usepackage{dsfont}
\usepackage{hyperref}
\usepackage{tablefootnote}

\usepackage{times}
\usepackage{latexsym}
\usepackage{comment}
\usepackage{amsmath}
\usepackage{amssymb}
\usepackage{float}
\usepackage{booktabs}
\usepackage[T1]{fontenc}
\usepackage[utf8]{inputenc}
\usepackage{microtype}
\usepackage{inconsolata}
\usepackage{graphicx}

\title{ M-Wanda: Improving One-Shot Pruning for Multilingual LLMs}

\author{Rochelle Choenni$^1$ \and Ivan Titov$^{1,2}$ \\
  University of Amsterdam$^1$ \\
  University of Edinburgh$^2$\\
  \texttt{r.m.v.k.choenni@uva.nl}, \texttt{ititov@inf.ed.ac.uk} \\}

\begin{document}
\maketitle

\begin{abstract}
Multilingual LLM performance is often critically dependent on model size. With an eye on efficiency, this has led to a surge in interest in one-shot pruning methods that retain the benefits of large-scale pretraining while shrinking the model size. However, as pruning tends to come with performance loss, it is important to understand the trade-offs between multilinguality and sparsification. In this work, we study multilingual performance under different sparsity constraints and show that moderate ratios already substantially harm performance. To help bridge this gap, we propose M-Wanda, a pruning method that models cross-lingual variation by incorporating language-aware activation statistics into its pruning criterion and dynamically adjusts layerwise sparsity based on cross-lingual importance. We show that M-Wanda consistently improves performance at minimal additional costs. We are the first to explicitly optimize pruning to retain multilingual performance, and hope to inspire future advances in multilingual pruning.\footnote{\url{https://github.com/RochelleChoenni/M-Wanda}.}
\end{abstract}

\section{Introduction}

Large language models (LLMs) have demonstrated strong multilingual capabilities, with their ability to process and generate text in numerous languages improving substantially as the model size increases~\citep{he2024scaling}. This emergent multilingualism can largely be attributed to the vast amount of multilingual data used for pretraining and the increased model capacity that allows for better generalization over linguistic patterns across multiple languages. However, the steep increase in model scale comes with substantial computational and environmental costs, making efficient deployment in resource-constrained environments challenging~\citep{ogueji2022intriguing}. To address these challenges, model compression techniques, such as pruning, quantization, and distillation, have been widely explored to reduce model size while retaining performance~\citep{zhu2024survey}.
However, despite the effectiveness of such methods, their evaluation focuses mainly on maintaining English performance~\citep{yangllmcbench}, with limited consideration of their impact on multilingual performance~\citep{zeng2024multilingual, kurz2024investigating, ogueji2022intriguing}. Given that multilingual performance is crucial for equitable LLMs, ensuring that compression does not disproportionately harm performance in non-English languages is essential.

In this paper, we study the effect of sparsity on the multilingual performance of six open-source LLMs of varying sizes. For model compression, we focus on a SOTA one-shot unstructured pruning method-- Wanda~\cite{sun2023simple}-- and evaluate language modeling abilities and zero-shot task performance at varying sparsity levels across 15 languages and six downstream tasks. Our results show that Wanda, despite its strong performance in English, causes substantial degradation in multilingual performance, particularly at sparsity levels higher than 50$\%$ and in underrepresented languages. These findings highlight an important limitation. As Wanda was developed to optimize for global importance, we hypothesize that it fails to account for cross-lingual variation in neuron importance, despite being exposed to multilingual calibration data, which leads to the removal of weights that are important for specific languages.

To help bridge this gap, we propose a novel multilingual pruning method, M-Wanda, which is a multilingual extension of Wanda. M-Wanda improves on Wanda by incorporating language-aware input activation statistics to better inform pruning decisions at minimal additional costs. Moreover, M-Wanda dynamically adjusts sparsity ratios across layers based on cross-lingual correlation scores, ensuring that layers that are important for cross-lingual sharing are pruned less aggressively. Together, these techniques allow us to better balance the contribution of shared and specialized neurons to weight importance. To the best of our knowledge, our work is the first to optimize pruning for multilingual retention, and to explicitly model cross-lingual activation variance and inter-language correlation to guide pruning decisions and layerwise sparsity allocation.

We show that M-Wanda consistently reduces perplexity across all languages and that this translates into performance improvements on all downstream tasks. Importantly, we show that M-Wanda generalizes well beyond the set of languages included in the calibration data. In addition, we show that the techniques introduced in M-Wanda can also be integrated with RIA~\citep{zhang2024plug}, a more recent pruning method, thus showcasing their general usefulness in extending pruning to a multilingual setting. 
Finally, our findings highlight the need to evaluate pruning methods beyond English-centric compression benchmarks~\citep{yangllmcbench} and emphasize the importance of optimizing the pruning strategy to preserve multilingual performance. In doing so, we hope to contribute to the development of more efficient LLMs that remain effective in many languages.

\section{Background and related work}

\subsection{Compression through pruning}

Pruning reduces model size by removing unnecessary weights or neurons~\citep{lecun1989optimal} and can be grouped into iterative and one-shot methods. Iterative methods~\citep{frankle2018lottery, blalock2020state} repeatedly prune a small percentage of weights, followed by retraining to recover performance, until a target threshold is met. While this does not require a predefined sparsity ratio, the additional training cycles can be expensive. One-shot methods, instead, remove a predefined fraction of weights in a single pass after the model is trained to convergence, and do not require retraining~\citep{frantar2023sparsegpt, sun2023simple}. It has gained popularity because of its simplicity and ability to maintain competitive performance.

\subsection{One-shot pruning methods} 
SparseGPT~\citep{frantar2023sparsegpt} introduced one-shot pruning by sequentially processing model layers and solving a local quadratic optimization problem to minimize reconstruction error under sparsity constraints. Yet, this requires a weight update after pruning and backward passes for gradient computation. \citet{sun2023simple} show that SparseGPT can be simplified to a gradient-free variant that achieves competitive performance without the need for parameter updates, i.e. Wanda.
\paragraph{Wanda method}
To determine the importance of model weights, \citet{sun2023simple} propose to incorporate the absolute weight value and the norm of the input activations of the neurons into the pruning criterion. Formally, let the input to a layer be denoted by  $X \in \mathbb{R}^{(N \times T) \times C_{\text{in}}}$, where $N$ is the batch size and $T$ the sequence length. The weight matrix \( W \in \mathbb{R}^{C_{\text{out}} \times C_{\text{in}}} \) connects the input features to the output units. For each weight element \( W_{i,j} \), the importance score is defined as:
\begin{equation}\label{eq:wanda}
  \mathbf{S}_{i,j} = |W_{i,j}| \cdot \|X_{j}\|_{2}   
\end{equation}
where $\|X_j\|_2$ denotes the $\ell_2$-norm of the $j$-th input feature column across all $N \times T$ tokens. This score reflects the contribution of each weight based on both its magnitude and the aggregated strength of the corresponding input feature. Note that, collecting input activation statistics requires a set of input samples which we refer to as \emph{calibration data}. Finally, a strong commonly used baseline is magnitude pruning~\citep{han2015learning}, in which only the weight magnitude: $\mathbf{S}_{i,j} = |W_{i,j}|$ is considered.

\paragraph{Layerwise sparsity}\label{sec:owl}
Sparsity allocation methods were developed to mitigate model degradation by enforcing different sparsity ratios across layers, and have been shown to improve performance~\citep{li2024discovering, huang2025determining}.
Rather than pruning uniformly, such methods estimate how much to prune based on the layers' sensitivity or redundancy. Concretely, given a global sparsity ratio $R$, the goal is to derive a set of target layerwise sparsity ratios [$r_0$, $r_1$, .., $r_L$] such that: $\frac{1}{L+1}\sum^{L}_{n=0}r_n = R$. 

One such method is Outlier Weighted Layerwise sparsity (OWL)~\citep{yin2024outlier}, which uses per-layer outlier counts (i.e. activations that exceed $M$ times the mean) as global importance scores $C=$ [$c_0$, $c_1$, .. ,$c_L$] for allocating sparsity. To prevent extreme imbalances between layers, they introduce a hyperparameter $\gamma$ that restricts each ratio to fall within a small interval around the global sparsity rate, specifically $r_n \in$  [$R - \gamma, R + \gamma$], while maintaining a mean sparsity ratio of $R$ across all layers.  To achieve this, the raw importance scores $C$ are rescaled to the range [$0$, $2\lambda$] and shifted so that the resulting values are centered around $R$. Following the intuition that layers that are more important should be pruned less, the sparsity ratios are then defined as: $r_n = 1 - c_n$.
\subsection{Multilingual pruning}

\citet{ogueji2022intriguing} first studied the effect of model pruning on multilingual performance. 
However, their scope was limited to iterative methods, smaller models, and a single task. More recently, \citet{zeng2024multilingual, kurz2024investigating} studied multilingual performance of LLMs using SparseGPT and Wanda. They both study how varying the composition of calibration data from different languages affect performance, and show that using a mixture of languages yields better results. However, both studies are limited to modifying the calibration data, without altering the pruning method itself, and restricting analysis to compression at 50\%. In this work, we show that 50\% sparsity already substantially harms multilingual performance. Moreover, this sparsity ratio is enforced uniformly across model layers, despite substantial evidence that model layers play different roles in language-specific and cross-lingual processing~\citep{tang2024language, kojima2024multilingual}. 
We, instead, study multilingual performance under different sparsity constraints and introduce M-Wanda, a novel pruning method that builds on Wanda by using multilingual calibration data, incorporating language-aware scoring, and combining it with an OWL-inspired dynamic sparsity allocation method.

\section{M-Wanda method}\label{sec:wanda}

While \citet{zeng2024multilingual, kurz2024investigating} show that using Wanda with multilingual calibration data improves performance, Wanda was developed to preserve weights that are globally important, and by averaging input activations across languages, we might suppress language-specific signals that are essential for multilingual retention. Thus, we enhance Wanda in three key ways: (1) We assign \textbf{\emph{layerwise sparsity}} based on the degree of cross-lingual activation similarity, applying less aggressive pruning to layers that are more important for cross-lingual sharing. (2) We incorporate \textbf{\emph{cross-lingual activation variance}} into the pruning criterion to encourage retention of specialized neurons that might, for instance, support underrepresented or typologically distinct languages. (3) We introduce an \textbf{\emph{activation probability}} term to discourage retention of high-variance neurons that rarely activate, helping to filter out noisy or spurious features. Together, these additions bias pruning toward preserving both shared and consistently active specialized neurons, thereby improving multilingual retention at minimal additional costs.

\subsection{Correlation Weighted Layerwise (CWL) sparsity}

We introduce \emph{Correlation Weighted Layerwise (CWL) sparsity} to guide sparsity allocation decisions across model layers. In contrast to OWL~\citep{yin2024outlier}, which scores layer importance based on outlier counts, CWL uses Pearson correlation coefficients to approximate activation similarity both across and between languages to determine importance. We hypothesize that layers that exhibit high inter-language activation similarity are more involved in cross-lingual sharing and better facilitate multilingual generalization. As such, we apply less aggressive pruning to them. However, when intra-language correlation scores are low, this suggests instable or noisy representations. To correct for this, we adjust the inter-language correlation score using intra-language scores. 

Concretely, we first compute Pearson correlation scores between the mean input activation (aggregated across tokens) for each language and sublayer $\mu_{\ell}^{(k)}$ of the attention or MLP block.\footnote{Note that input activations are shared between query, key and value, and between the MLP gate and up projection layers.}
To compute the average inter-language correlation score for sublayer $k$ and a set of languages $\mathcal{L}$, we then take the mean of all pairwise correlations:
\begin{equation}
    \text{Inter}^{(k)} = \frac{2}{|\mathcal{L}|(|\mathcal{L}| - 1)} \sum_{i < j} \text{corr}(\mu_{\ell_i}^{(k)}, \mu_{\ell_j}^{(k)})
\end{equation}
\noindent Moreover, we adjust inter-language scores using intra-language scores, by assigning more importance when both are high, yielding: 
\begin{equation}
    c^{(k)} = \text{Inter}^{(k)} \cdot \sum_{\ell\in \mathcal{L}}\text{Intra}_{\ell}^{(k)}
\end{equation}
\noindent This score reflects how shared representations are between languages and how stable they are within languages.
To obtain a single importance score for each layer $n$: [$c_0$, $c_1$, .. $c_L$], we take the average over all sublayers. We then apply the same procedure as OWL, described in Section~\ref{sec:owl}, to ensure that the mean sparsity is equal to the global ratio $R$, and assign layers with more importance, lower ratios: $r_n = 1 - c_n$. We find that setting $\lambda$ to $0.04$ generally works well across LLMs.

\subsection{Cross-lingual activation variance}  
Recall from Eq~\ref{eq:wanda} that Wanda incorporates both weight importance and activation strength. We now enhance the activation scores by storing the mean of activation values per language $\mu_{\ell}$ and computing the variance in neuron activation across languages:
\begin{equation}
    \text{Var}_{inter} = \frac{1}{|\mathcal{L}|} \sum_{\ell \in \mathcal{L}} \left( \mu_{\ell} - \bar{\mu} \right)^2
\end{equation}
\noindent By adding this inter-language variance score, we give more importance to neurons whose input activations show highly variable responses across languages, meaning that they might be very important to \textit{some} specific languages. 

Yet, the input activations within a language also fluctuate between input samples. If the intra-language variance is high, it introduces noise into our pruning metric, making the inter-language variance less reliable. Therefore, we assess how much neuron activation varies between languages relative to how much it varies within individual languages:

\begin{equation}\label{eq:inter-norm}
    VAR = \frac{\mathrm{\text{Var}}_{inter}}{\frac{1}{|\mathcal{L}|} \sum^{\mathcal{L}}_{\ell=1} \mathrm{\text{Var}}^{\ell}_{intra}}
\end{equation}
\noindent This means that we assign higher scores to neurons that exhibit high inter-language variance but low intra-language variance.
\begin{equation}\label{eq:var}
    A_{X_{j}} =  \|X_{j}\|_{2} + \lambda \cdot  VAR
\end{equation}
To balance the trade-off between language-specificity and generalization, we add a scaling term $\lambda$ for which the optimal value is found through a grid search. Also, note that before adding variance scores, we apply min-max normalization.

\subsection{Activation probability} 
Finally, we correct the overall weight importance scores based on the average activation probability across languages. This is motivated by the idea that high-variance neurons that are upweighted by Eq.~\ref{eq:var}, but rarely activate, are noisy and should be filtered out. To compute this activation probability, we simply count how many times the input activations are higher than some threshold value $\epsilon$. Given that recent LLMs rely on activation functions that also allow for negative activation that can contain meaningful information, we consider absolute activation values.\footnote{Using absolute values was found to work better than positive ones, showing that negative signals carry information.} As such, we end up with the following pruning metric:
\begin{equation}
  \mathbf{S}_{i,j} = (|W_{i,j}| \cdot A_{X_{j}}) \cdot P(\mathds{I}(abs(X_{j}) > \epsilon))   
\end{equation}
where $\mathds{I}(\cdot)$ is the indicator function.

\section{Experiments}

\paragraph{Calibration and test languages}
For calibration and evaluation we use 15 languages: English, German, Spanish, French, Italian, Portuguese, Hindi, Russian, Korean, Japanese, Vietnamese, Chinese, Indonesian, Turkish, Arabic. These languages belong to 8 different families, 11 sub-families and cover 7 writing scripts.

\paragraph{Calibration data}
Following prior work, we use 128 random samples of 2048 tokens from the mulitlingual C4 (MC4) dataset for calibration~\citep{raffel2020exploring}. While recent studies show that the data source affects pruning quality~\citep{williams2024impact, ji2024beware, bandari2024c4}, we use MC4 to limit the scope of this work and ensure comparability with existing literature. To adhere to the 128 samples maximum, our calibration data includes 16 samples from English and 8 from all other test languages.

\begin{figure*}[!t]
    \centering
    \scalebox{0.8}{
    \includegraphics[width=\linewidth]{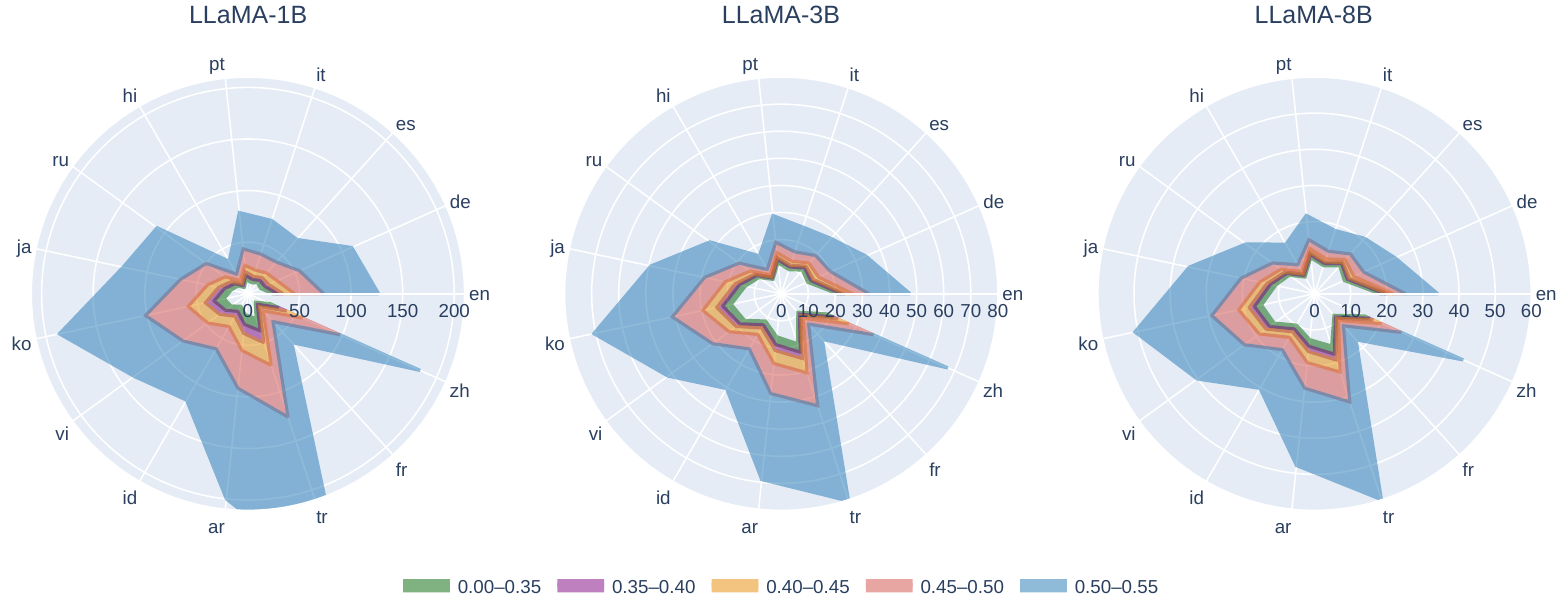}}
    \caption{The effect of Wanda pruning under different sparsity ratios on the perplexity of each calibration language. Colored areas denote the increase in perplexity when increasing the sparsity ratio. Note that the perplexity scores are on different scales across models.} 
    \label{fig:ppl_wanda}
\end{figure*}

\begin{figure*}
    \centering
\includegraphics[width=\linewidth, height=3.5cm]{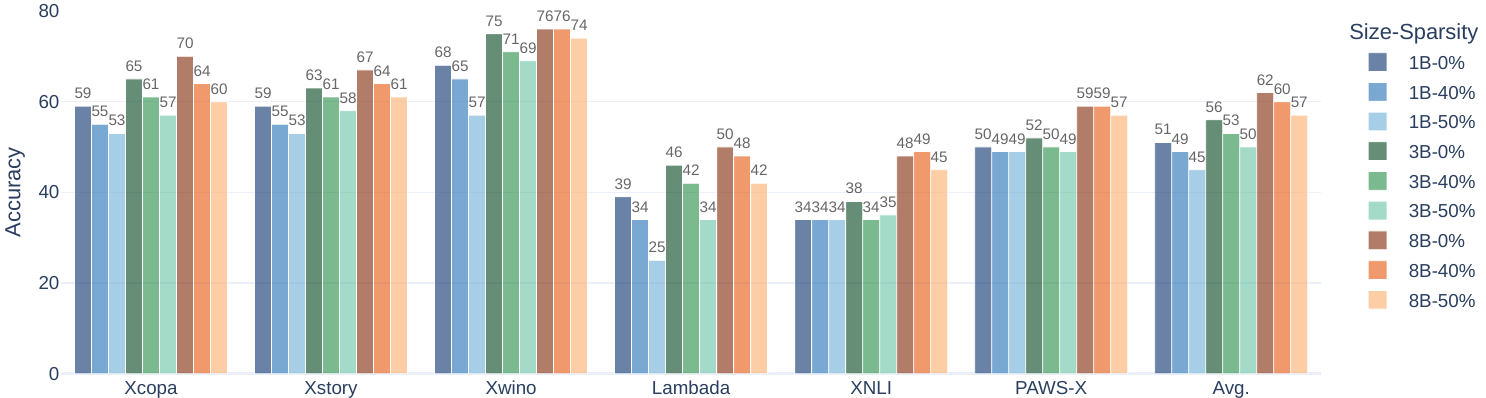}
    \caption{Performance in accuracy ($\%$) given different sparsity ratios used on different sizes of Llama3. Zero-shot results are averaged across test languages per downstream task. }
    \label{fig:downstream}
\end{figure*}

\paragraph{Models}
We study six open-source LLMs at different model sizes:
Llama3 (1B, 3B and 8B)~\citep{grattafiori2024llama}, Aya-23 (8B)~\citep{dang2024aya}, OLMo-7B~\citep{Groeneveld2023OLMo} and Bloomz-7b1~\citep{muennighoff2023crosslingual}.

\paragraph{Zero-shot performance} We perform zero-shot evaluation on six tasks that test the LLMs ability on reasoning (Xstorycloze, Xcopa)~\citep{ponti2020xcopa}, coreference resolution (Xwinograd)~\citep{muennighoff2023crosslingual}, reading comprehension (Lambada)~\citep{paperno2016lambada}, natural language understanding (XNLI)~\citep{conneau2018xnli}, and paraphrasing (PAWS-X)~\citep{yang2019paws}. For consistent evaluation we employ the eleuther-AI evaluation harness.\footnote{\url{https://github.com/EleutherAI/lm-evaluation-harness}}

\paragraph{Model perplexity}
We test general language modeling abilities by measuring perplexity on datasets different from the one used for calibration. Specifically, we evaluate perplexity on the entire Flores-101 (\texttt{dev}+\texttt{devtest}) dataset which contains parallel data from Wikipedia. To test whether our results are robust across different domains, we also evaluate on the XL-Sum dataset that contains high-quality articles from BBC~\citep{hasan2021xl}.

\section{Results}
In Section~\ref{sec:baseline}, we first show how pruning under different sparsity constraints affects multilingual LLMs of different sizes. Motivated by these findings, we show in Section~\ref{sec:mwanda-res} how our M-Wanda method can help mitigate some of the multilingual performance loss induced by pruning.

\subsection{Wanda's impact on multilinguality}\label{sec:baseline}

We prune our models using Wanda with sparsity ratios between 35 and 60$\%$ at 5 percent intervals. In Figure~\ref{fig:ppl_wanda}, we see that across all languages and different sizes of Llama, the perplexity has already substantially increased when going from 45 to 50$\%$ sparsity (red area), especially on underrepresented languages (typically not from the Indo-European family). This sheds doubt on the common practice of adopting the default sparsity ratio of 50$\%$ in the multilingual setting~\citep{zeng2024multilingual, kurz2024investigating}. Importantly, this same degradation is not found in English when only using English calibration data (see Appendix~\ref{app:english}), the setting used in the original paper~\citep{sun2023simple}. 

Similarly, a clear degradation across all downstream tasks is visible when going from 40 to 50$\%$ sparsity; see Figure~\ref{fig:downstream}. In fact, when studying how larger models pruned to 50$\%$ of their original capacity compared to their smaller dense counterparts (i.e. Llama 3B at 50$\%$ versus Llama 1B and Llama 8B at 50$\%$ versus Llama 3B), we see that they are not able to outperform them despite still having a larger capacity. 

\begin{table*}[!t]
    \centering
    \begin{tabular}{l|llllll}
    \toprule
        Method & Llama3-1B & Llama3-3B & Llama3-8B & Aya-23-8B & Bloomz-7b1 & OLMo-7B\\
        \midrule
        Magnitude & 17605 & 1579 & 403 & 36.12 & 29.64 & 33.55\\
        RIA$^*$ & 71.75 & 27.88 & 20.45 & 25.28 & \textbf{24.05}&30.45 \\
        Wanda     & 63.29 & 26.42 & 19.59 & 24.34 & 24.71 & 23.23\\
        M-Wanda   & \textbf{59.56} (6$\%$$\downarrow$) & \textbf{24.56} (7$\%$$\downarrow$) & \textbf{18.57} (5$\%$$\downarrow$) & \textbf{23.87} (2$\%$$\downarrow$) & 24.32 (2$\%$$\downarrow$) & \textbf{21.54} (7$\%$$\downarrow$)\\
        \bottomrule
    \end{tabular}
    \caption{Average perplexity on Flores across all calibration languages at a sparsity ratio of 50$\%$. For M-Wanda, we also report the relative percentage decrease compared to Wanda.$^*$Refer to Section~\ref{sec:ria} for an introduction to RIA. }
    \label{tab:main_avg_ppl}
\end{table*}

\begin{table*}[!t]
    \centering
    \begin{tabular}{c|cccccc|c}
    \toprule
        & Xcopa & Xstory & Xwino & Lambada & XNLI & PAWS-X & Avg. \\
        \hline
        Wanda & 60.36 & 60.86 & 73.81 & 42.08 & 45.07 & 57.43 & 56.60 \\
        M-Wanda & \textbf{61.16} & \textbf{61.49} & \textbf{74.54} & \textbf{44.68} & \textbf{46.51} & \textbf{58.23} & \textbf{57.77} \\
        \bottomrule
    \end{tabular}
    \caption{Average performance (\%) on downstream tasks when using Wanda versus M-Wanda on Llama-8B.}
    \label{tab:avg_task}
\end{table*}

\subsection{Improvements with M-Wanda}\label{sec:mwanda-res}
In Section~\ref{sec:baseline}, we show that Wanda with 50\% sparsity leads to a substantial drop in multilingual performance. This degradation highlights an area of potential improvement for M-Wanda, and we hypothesize that more optimally balancing the importance between specialized and shared neurons would allow us to better retain multilingual performance. In Table~\ref{tab:main_avg_ppl}, we show how M-Wanda is able to reduce the average perplexity across languages for all models on the Flores dataset (see Appendix~\ref{app:hs} for the optimal hyperparameters selected for each model and Appendix~\ref{app:xlsum} for results on XL-Sum).\footnote{Perplexity from magnitude pruning is notably higher on Llama. We find that performance is reasonable in English, yet explodes on other languages, yielding high average scores.} Moreover, we find that this holds across different model sizes. Importantly, while lower perplexity does not always guarantee better performance on downstream tasks, in Table~\ref{tab:avg_task} we show that the improvements achieved by M-Wanda are substantial enough to improve performance in all six downstream tasks.

\begin{figure}[!t]
    \centering
    \vspace{-0.6cm}
    \scalebox{0.72}{\includegraphics[width=\linewidth]{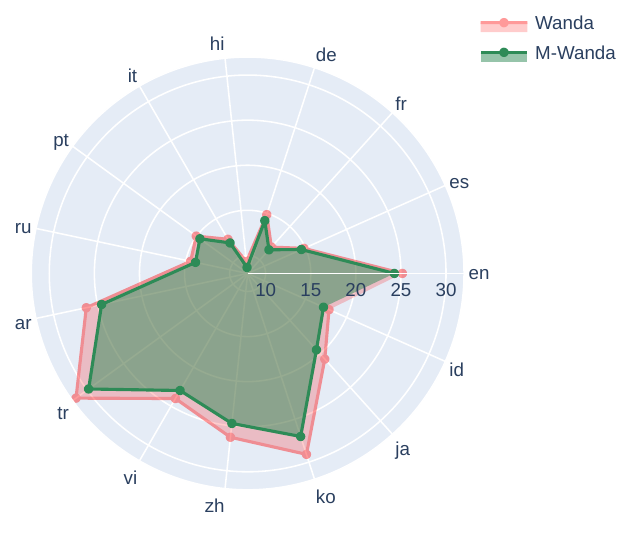}}
    \vspace{-0.5cm}
    \caption{Perplexity scores per language from Llama-8B pruned using Wanda versus M-Wanda.}
    \label{fig:abs_ppl_wanda}
\end{figure}

When taking a closer look at the effect of M-Wanda on individual test languages in Figure~\ref{fig:abs_ppl_wanda}, we see that M-Wanda consistently reduces the perplexity on all 15 languages for Llama-8B. Moreover, we see that languages most typologically different from English, i.e. Arabic, Turkish, Vietnamese, Chinese, Korean and Japanese, obtain larger absolute gains from M-Wanda than the Indo-European languages. Similarly, when looking at downstream task improvements, we find that M-Wanda tends to consistently improve performance on all individual test languages, with a few exceptions (mostly English), see Appendix~\ref{app:downstream} for full results. 

\subsubsection{Generalization to unseen languages} 
\begin{figure*}[!t]
    \centering
    \scalebox{0.9}{\includegraphics[height=5cm, width=\linewidth]{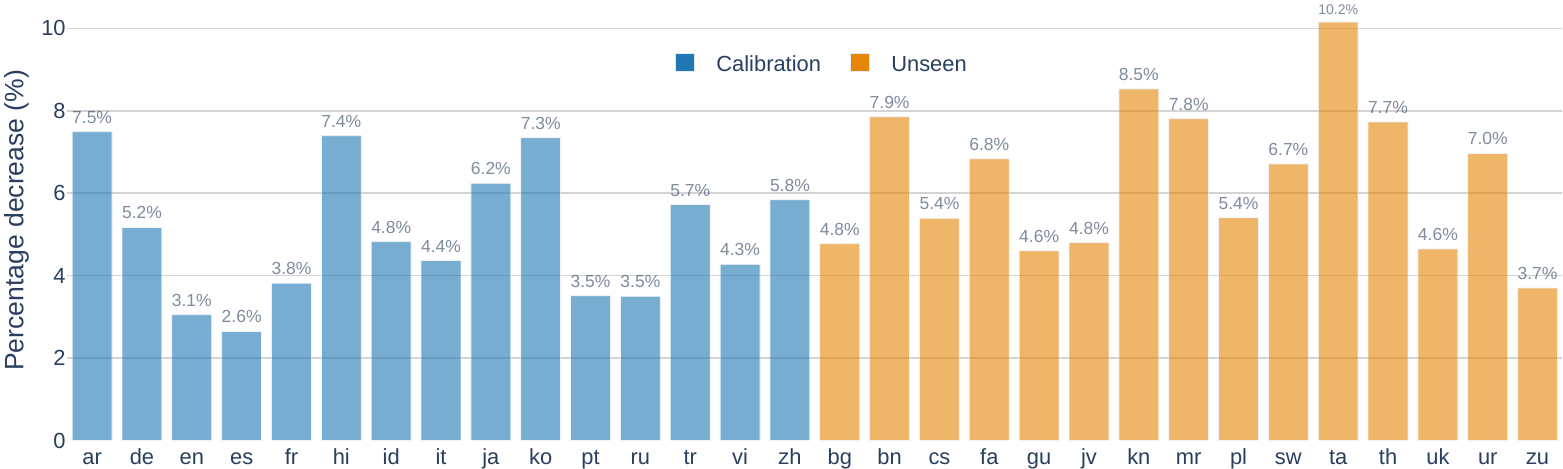}}
    \caption{Relative percentage decrease in perplexity when using M-Wanda compared to Wanda for all 15 calibration and 15 unseen test languages. Results are reported for Llama-8B.}
    \label{fig:relative}
\end{figure*}

Previously, we used the same set of languages for calibration and testing. We now test whether the performance improvements also generalize beyond our calibration languages. As such, we use 15 different languages for evaluation: Czech, Polish, Ukranian, Bulgarian, Tamil, Marathi, Urdu, Bengali, Kannada, Gujarati, Javanese, Thai, Swahili, Zulu, and Persian. In Figure~\ref{fig:relative}, we show that our method consistently reduces perplexity across all languages, despite them not being included in the calibration set. Specifically, M-Wanda results in a 6$\%$ decrease in average perplexity compared to Wanda (18.98 versus 20.09), which is higher than on the calibration languages itself. This is likely because our unseen languages include many more underrepresented languages, and our method seems to book larger performance gains on those. Importantly, however, this suggests that M-Wanda more generally helps to preserve language variance and is not only adjusting to the language-specific patterns of the calibration languages.

\subsubsection{Effectiveness at different sparsity levels}  
\begin{figure}[!t]
    \centering
    \scalebox{1}{\includegraphics[width=\linewidth]{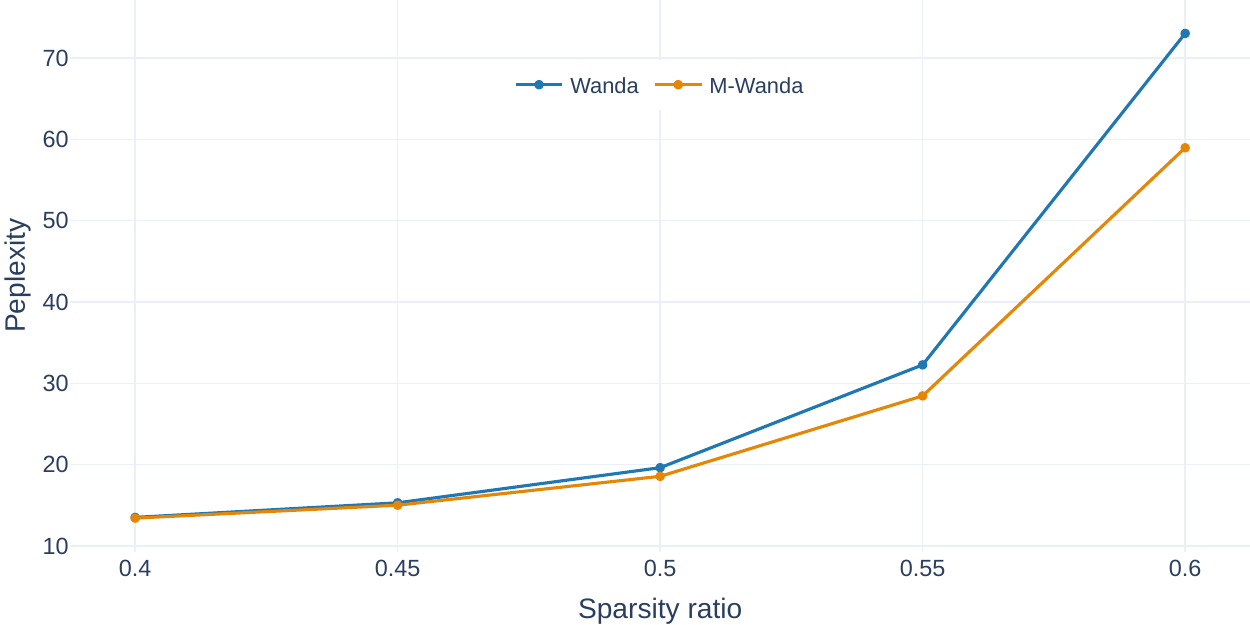}}
    \caption{Average perplexity scores across languages as an effect of higher sparsity ratios when applying Wanda and M-Wanda to Llama-8B. }
    \label{fig:pruning_ratios}
\end{figure}

While we already saw that M-Wanda improves performance across different model sizes, we now also test its effectiveness across different sparsity ratios. In Figure~\ref{fig:pruning_ratios}, we plot the average perplexity scores obtained with Wanda and M-Wanda at different sparsity levels. We see that M-Wanda remains effective at higher ratios and that the average improvement of M-Wanda over Wanda increases substantially when applying more aggressive pruning. At the extreme sparsity ratio of 70$\%$ we find that M-Wanda reduces average perplexity by as much as 52$\%$ (see Appendix~\ref{app:ratios} for full results).

\subsubsection{Robustness analysis}

\paragraph{Sensitivity to calibration samples} While we limited the scope of this paper to randomly selecting calibration samples from the MC4 dataset, we now also test the sensitivity of our method to the calibration set. Specifically, we use 3 random seeds to select calibration data and recompute average perplexity using  Wanda versus M-Wanda. We find that across all three runs, M-Wanda outperforms Wanda. On average Wanda obtains a perplexity of 19.37$\pm 0.27$ and M-Wanda 18.63$\pm 0.22$.

\begin{figure}[!t]
    \centering  \includegraphics[width=\linewidth]{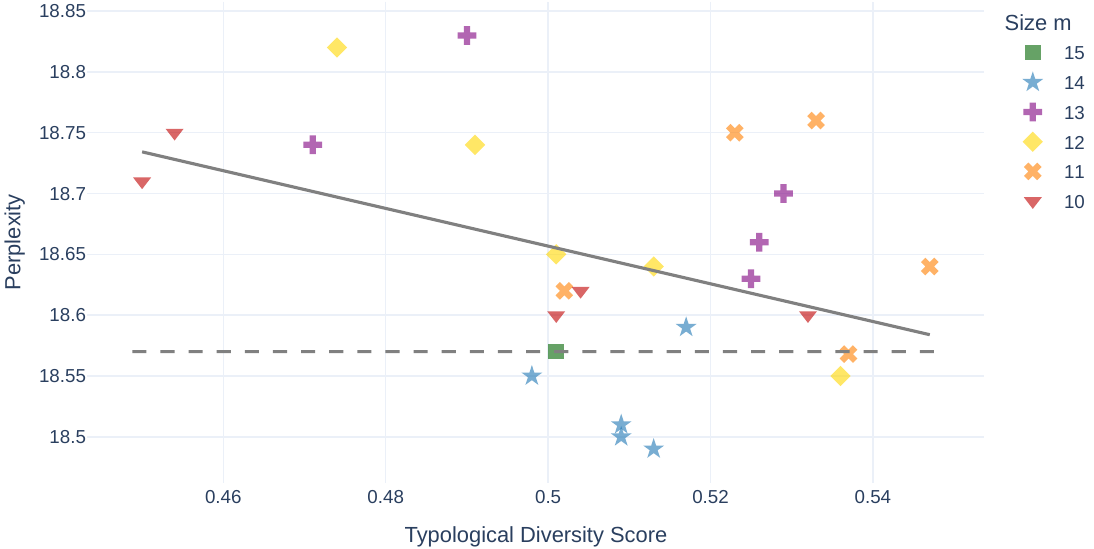}
    \caption{Average M-Wanda perplexity on Llama-8B as a function of the typological coverage of the calibration languages. Subsets are colored based on their size $m$.}
    \label{fig:tradeoff}
\end{figure}

\paragraph{Sensitivity to calibration languages}
Finally, we now study how selecting different subsets of languages from the full calibration set affects performance. These subsets vary both in size, which impacts the number of samples per language and, consequently, the robustness of language-specific signals, and in their typological composition, which might influence how well calibration generalizes across languages.
To study this, we draw multiple random subsets of languages for calibration. Specifically, we each time sample 5 unique subsets of size $m$ uniformly at random from the set of all languages $\mathcal{L}$:
$\mathcal{S}_m^{(i)}\sim\text{Unif}\left( \left\{ S \subset \mathcal{L} \,:\, |S| = m \right\} \right)$.
In Figure~\ref{fig:tradeoff}, we plot the average perplexity on the full calibration set $\mathcal{L}$ as a function of the typological diversity of the calibration languages in the different language subsets of size $m$. These diversity scores are defined as the mean of pairwise cosine similarity between their URIEL language representations~\citep{malaviya2017learning}.\footnote{We use \texttt{syntax\_knn} features from the \href{https://github.com/antonisa/lang2vec}{\texttt{Lang2Vec}} library.} In general, we find that higher typological diversity leads to better performance. However, we also observe that a few language subsets can outperform the full calibration set, suggesting that optimal calibration may depend more on carefully selecting which languages are chosen than on increasing its size. 

\section{Ablation study}

To understand where the performance improvements of M-Wanda come from, we now perform an ablation study, isolating the impact of individual enhancements that were added to the original Wanda method.
\begin{figure}[!t]
    \centering  \includegraphics[width=\linewidth]{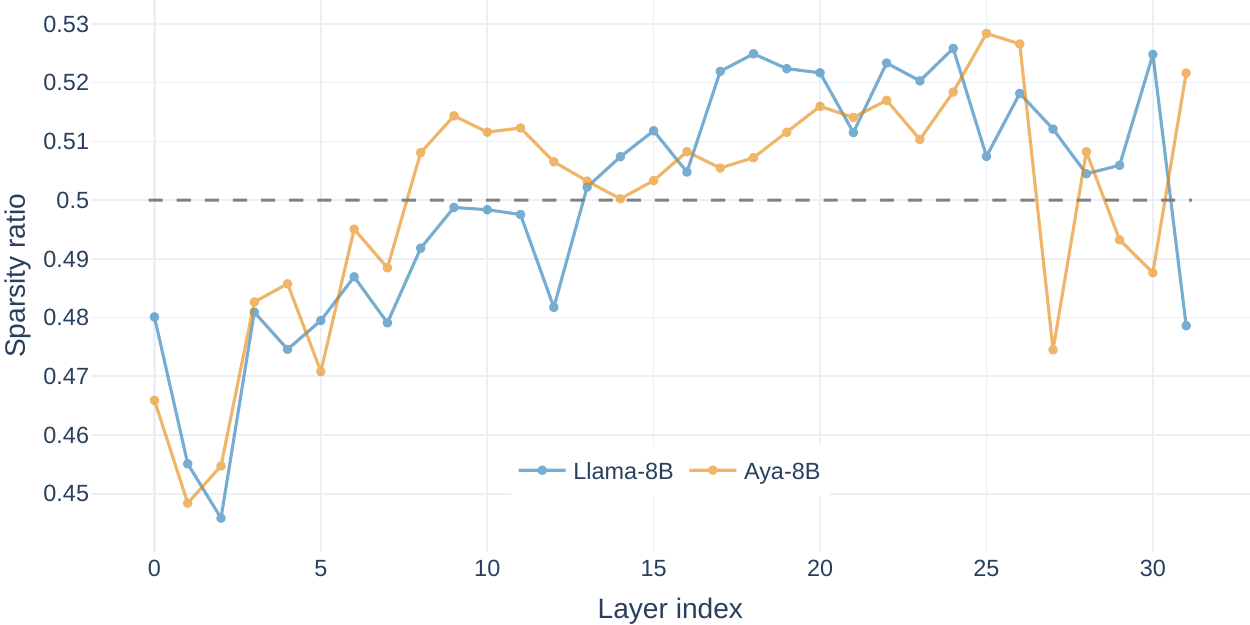}
    \caption{Layerwise sparsity allocation using CWL.}  \label{fig:layerwise_sparsity}
\end{figure}
When we combine the original Wanda metric with OWL instead of CWL allocation, we find that OWL\footnote{We used optimal hyperparameters reported in the original paper i.e., $M \in [3,5]$ and $\gamma=0.08$, but found $\gamma=0.04$, as used for CWL, works better and thus report scores using the latter for a more fair comparison.}  reduces average perplexity to a lesser extent and for the 1B model even worsens it, see Table~\ref{tab:ablation}. 
 Importantly, we also find that enhancing $\text{Wanda}_{\textbf{+OWL}}$ with cross-lingual variation and activation probability further improves performance to 19.09 on LLama-8B. This shows that our proposed enhancements can more generally be paired with different allocation methods and do not work exclusively in combination with CWL. 
 
 Moreover, in Figure~\ref{fig:layerwise_sparsity} we plot the sparsity ratio allocated per layer for Llama-8B and Aya-8B using CWL. In general, we see that the lower layers and the last few top layers receive less aggressive pruning. The fact that this results in better multilingual performance can likely be connected to the fact that these layers have been shown to be more involved in cross-lingual processing~\citep{zhaolarge} (see Appendix~\ref{app:owl} for allocation results using OWL). 

\begin{table}[!t]
    \centering
    \setlength{\tabcolsep}{2pt}
    \begin{tabular}{lcccc}
    \toprule
         & Wanda & Wanda$_{\textbf{+OWL}}$& Wanda$_{\textbf{+CWL}}$ & M-Wanda \\
         \midrule
          1B & 63.29 & 65.10 & 60.50  & \textbf{59.56}\\ 
           3B &26.42 & 26.02 & 24.61  & \textbf{24.56}\\ 
         8B &19.59 &  19.14& 18.61 & \textbf{18.57}\\
         \bottomrule
    \end{tabular}
    \caption{M-Wanda ablation on the Llama3 models. We report average perplexity  on the calibration languages.}
    \label{tab:ablation}
\end{table}

\section{Extendability to other pruning methods}\label{sec:ria} Relative Importance and Activations (RIA) is a SOTA pruning method that has been shown to outperform Wanda~\citep{zhang2024plug}. It aims to improve upon Wanda by re-evaluating the importance of each weight element $W_{ij}$ based on all connections that originate from the input neuron $i$ or lead to the output neuron $j$:
\begin{align}
     \mathbf{RIA}_{i, j} &=  \mathbf{RI}_{i, j} \cdot (\|X_{j}\|_{2})^{\alpha} \nonumber \\
     &= \left( \frac{|W_{i,j}|}{\sum |W_{*j}|} + \frac{|W_{i,j}|}{\sum |W_{i*}|} \right) \cdot (\|X_{j}\|_{2})^{\alpha}
\end{align}
\noindent where $\sum|W_{*j}|$ and $\sum|W_{*i}|$ sum over the absolute values of the weights in input channel $j$ and output channel $i$ respectively.
Yet, while we find that RIA outperforms Wanda on English at 50$\%$ sparsity (25.05 versus 25.16 on Llama-8B), the average perplexity across all 15 calibration languages tends to increase instead (20.45 versus 19.59 on Llama-8B). This further highlights the need for multilingual evaluation of pruning methods. Nonetheless, to test the compatibility of our proposed method with different pruning criterion, we now add cross-lingual variance and activation probability to RIA and apply CWL to obtain layerwise sparsity, yielding M-RIA:
\begin{equation}
\begin{aligned}
  \mathbf{S}_{i,j} &=  (\mathbf{RI}_{i,j} \cdot A_{X_{j}}) \cdot P(\mathds{I}(|X_{j}| > \epsilon)) \\
  \text{where} \quad A_{X_j} &= (\|X_{j}\|_{2})^{\alpha} + \lambda \cdot VAR
\end{aligned}
\end{equation}
\noindent Note that we adopt $\alpha$=0.5 which \citet{zhang2024plug} found to be optimal for various LLMs. In Table~\ref{tab:avg_ppl_ria} we show how M-RIA is also able to consistently improve over RIA, nicely demonstrating the general advantage of our proposed method for adaptation to a multilingual setting. 

\begin{table}[!t]
    \centering
    \begin{tabular}{l|ll}
    \toprule
        & RIA & M-RIA   \\ 
        \hline
        Llama3.2-1B & 71.75  & \textbf{66.22} (8$\%$$\downarrow$) \\
        Llama3.2-3B &  27.88 & \textbf{25.53} (8$\%$$\downarrow$)\\
         Llama3.1-8B& 20.45 & \textbf{19.03} (7$\%$$\downarrow$)\\ 
          Aya-23-8B & 25.28  & \textbf{24.82} (2$\%$$\downarrow$)\\ 
                   Bloomz-7b1 & 24.05 & \textbf{23.65} (2$\%$$\downarrow$) \\
             OLMo-7B & 30.45 & \textbf{26.21} (14$\%$$\downarrow$) \\
         \bottomrule
    \end{tabular}
    \caption{Average perplexity scores on the calibration languages for Flores using RIA ($\alpha$=0.5) at 50$\%$ sparsity.}
    \label{tab:avg_ppl_ria}
\end{table}

\section{Conclusion}
In this paper, we shed light on the limitations of SOTA pruning methods in a multilingual setting and introduce M-Wanda, a novel pruning method that explicitly models cross-lingual variation in weight importance. By incorporating language-aware activation statistics and adaptive sparsity allocation, M-Wanda substantially improves multilingual retention over existing methods, particularly for underrepresented languages and at high sparsity ratios. Our results show that multilingual pruning requires strategies that go beyond global importance scoring and highlight the benefits of considering the importance of specialized neurons. We hope that these insights help advance the state of multilingual pruning by underscoring the broader need for multilingual evaluation and design in LLM sparsification, and inspire new directions to improve multilingual pruning beyond the modification of calibration data. 

\section{Limitations}

Our improvements to the original Wanda method come at minimal additional computational costs. Specifically, we only compute additional statistics from the activation inputs that would already need to be collected for the original method. Moreover, we still use 128 calibration samples in total across all of our calibration languages. However, while we show that input activation statistics can help inform pruning decisions, unlike Wanda, M-Wanda does require tuning of the hyperparameters. To alleviate the need for manual tuning, future work could investigate how hyperparameters could automatically be adjusted based on the scale of the weights and activations. 

In addition, we limited the scope of this project to studying unstructured pruning, the setting for which Wanda was originally developed. However, \citet{sun2023simple} show that Wanda can also be extended to structured N:M pruning, which requires at most N out of every M contiguous weights to be non-zero (e.g. 2:4 or 4:8)~\citep{mishra2021accelerating}. While this usually results in lower performance, it is more amenable to practical inference speed-ups. Thus, future work should investigate how the core ideas behind M-Wanda generalize to structured pruning settings. 

\section*{Acknowledgement}
This work is supported by the Dutch National Science Foundation (NWO Vici VI.C.212.053).

\bibliographystyle{acl_natbib}

\appendix
\onecolumn

\section{Hyperparameter selection}\label{app:hs}

\begin{table}[H]
    \centering
    \begin{tabular}{l|ccccc}
    \toprule
        & $\lambda$ & $\epsilon$ & $\gamma$ & CWL block \\ 
        \hline
        Llama3-1B& 0.2 & 5e-5 & 0.04 & attn \\
        Llama3-3B& 0.02 & 1e-7 & 0.04 & attn \\
         Llama3-8B & 0.2 & 5e-5 & 0.04 & attn \\
         Aya-23-8B & 0.2 & 1e-7 & 0.04 & MLP \\
         OLMo-7B & 0.2 & 1e-7 & 0.04 & MLP\\
         Bloomz-7b1& 6 & 0 & 0.01 & attn \\
         
         \bottomrule
    \end{tabular}
    \caption{Optimal hyperparameters found for M-Wanda after a small search $\lambda \in [0.02, 0.2]$, $\epsilon \in [$5e-5, 1e-7$]$, $\gamma = [0.01, 0.04]$ and CWL block $\in [\text{attn}, \text{MLP}]$ .}
    \label{tab:hs}
\end{table}

 In Table~\ref{tab:hs}, we report the optimal hyperparameters used when applying M-Wanda to each model. Note that these are the hyperparameters used for both the results on the Flores dataset, reported in Table~\ref{tab:main_avg_ppl}, and the XL-Sum dataset, reported in Table~\ref{tab:avg_pp_xlsuml}. We find that, generally, $\lambda=0.2$ and $\gamma=0.04$ work well on most LLMs. 
 
 However, interestingly, the optimal hyperparameter value for $\lambda$ on Bloomz-7b1 is on a completely different scale, so we ran a different grid search for this model: $\lambda \in [6, 12]$ .  Moreover, we noticed that input activations from Bloomz-7b1 are sometimes equal to zero, resulting in suboptimal performance when applying the activation probabilities. Thus, while we found that this model benefits from cross-lingual variance and CWL, activation probability should be disabled ($\epsilon=0$) for the improvements reported in the main paper.

In addition, Wanda$_{\textbf{+OWL}}$  results reported in Table~\ref{tab:ablation}, required the tuning of the hyperparameter $M$. We found that for Llama 1B and 3B $M=3$ is optimal, yet for Llama 8B $M=5$ yields better results. Similarly, while $\gamma=0.08$ was reported to generally work best with OWL, we found that $\gamma=0.04$, as used for CWL, performed better. As such, those are the values used for the results reported in the table.

\section{XL-Sum results}~\label{app:xlsum}

\begin{table}[h]
    \centering
    \begin{tabular}{l|ll}
    \toprule
        & Wanda & M-Wanda   \\ 
        \hline
         Llama3.2-1B & 40.30 & \textbf{36.88} (8$\%\downarrow$) \\
        Llama3.2-3B &  16.40 & \textbf{15.61} (5$\%\downarrow$)  \\
         Llama3.1-8B& 12.18 & \textbf{11.57} (5$\%\downarrow$) \\
          Aya-23-8B & 15.46 & \textbf{15.14} (2$\%\downarrow$) \\
         Bloomz-7b1 & 20.35 &  \textbf{17.40} (14$\%\downarrow$) \\
       OLMo-7B & 15.28  & \textbf{14.34} (6$\%\downarrow$) \\
        \bottomrule
    \end{tabular}
    \caption{Average perplexity scores on the XL-Sum validation set across 13/15 languages at a sparsity ratio of 50$\%$. We use 500 articles for each language. Note: German and Italian are not covered by this dataset.}
    \label{tab:avg_pp_xlsuml}
\end{table}

\newpage

\section{Effectiveness of M-Wanda at different sparsity ratios}\label{app:ratios}
\begin{table}[h]
\centering
\begin{tabular}{lccccccccc}
\toprule
Sparsity & 30$\%$ & 35$\%$ & 40$\%$ & 45$\%$ & 50$\%$ & 55$\%$ & 60$\%$ & 65$\%$ & 70$\%$ \\
\midrule
Wanda   & 12.12 & 12.62 & 13.52 & 15.31 & 19.63 & 32.27 & 73.02 & 174.70 & 1743.14 \\
M-Wanda & 12.12 & \textbf{12.60} & \textbf{13.43} & \textbf{15.02} & \textbf{18.57} & \textbf{28.46} & \textbf{58.96} & \textbf{159.48} & \textbf{835.63} \\
\bottomrule
\end{tabular}
\caption{Average perplexity scores of Wanda and M-Wanda across different sparsity levels. For reference, the average performance of the full model is 11.38. Results are reported on Llama-8B.}
\end{table}

\section{English-centric pruning}
\label{app:english}
\begin{figure*}[h]
    \centering
    \includegraphics[width=\linewidth]{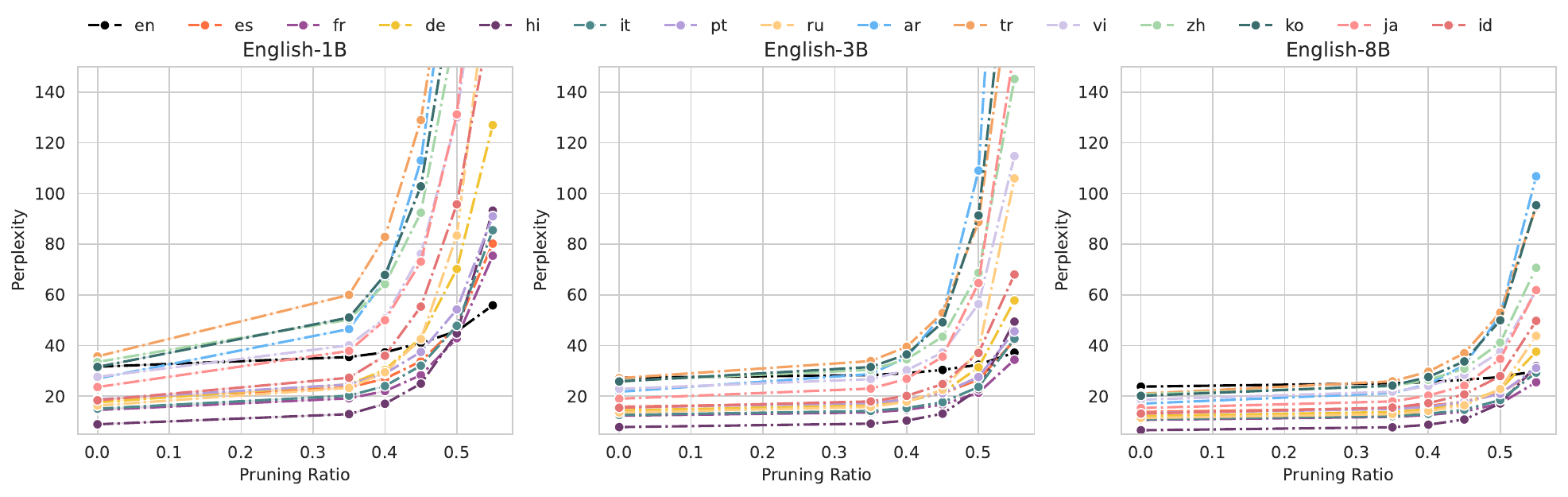}
    \caption{The effect of higher sparsity ratio's on the perplexity across languages. The calibration data is fully in English and perplexity is measured on the Flores dataset. Results are reported on the Llama3 models.}
    \label{fig:english-cali}
\end{figure*}

\section{Sparsity allocation with OWL}\label{app:owl}
\begin{figure}[h]  
\centering
\includegraphics[width=0.5\linewidth]{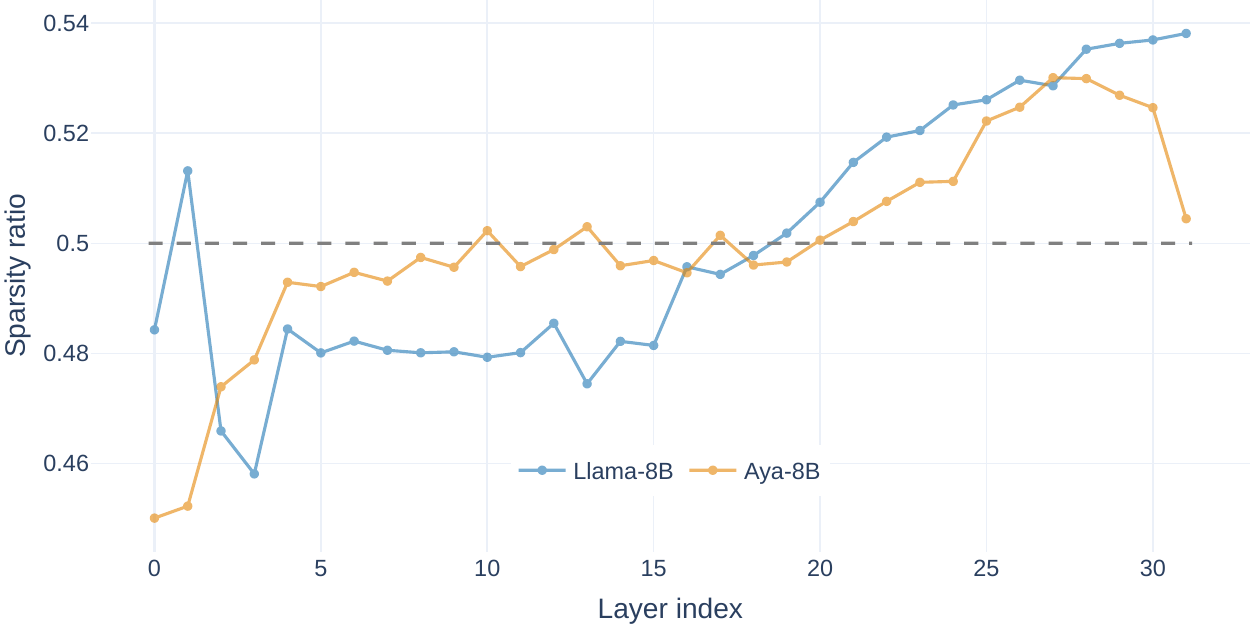}
    \caption{Layerwise sparsity allocation by OWL.}
    \label{fig:enter-label}
\end{figure}
\section{Downstream task results per language}~\label{app:downstream}

\begin{table*}[ht]
\centering
\begin{tabular}{l|lll}
\toprule
 & Lang. & Wanda & M-Wanda \\
\midrule
\multirow{5}{*}{XCOPA}
& id & 58.8 & \textbf{60.4} \\
& it & 62.8 & \textbf{63.2} \\
& tr & 57.8 & \textbf{58.6} \\
& vi & 59.6 & \textbf{61.0} \\
& zh & \textbf{62.8} & 62.6 \\
\hline
\multirow{7}{*}{XStory}
& ar & 52.9 & \textbf{53.8} \\
& en & \textbf{72.0} & 71.8 \\
& es & 65.0 & \textbf{65.2} \\
& hi & 56.8 & \textbf{57.8} \\
& id & 58.7 & \textbf{60.3} \\
& ru & 61.7 & \textbf{61.9} \\
& zh & 58.9 & \textbf{59.6} \\
\hline
\multirow{5}{*}{XWinograd}
& en & \textbf{86.7} & 86.0 \\
& fr & 69.9 & 69.9 \\
& pt & 73.8 & \textbf{74.1} \\
& ru & 66.7 & \textbf{68.9} \\
& zh & 72.0 & \textbf{73.8} \\
\hline
\multirow{5}{*}{Lambada}
& de & 34.4 & \textbf{36.3} \\
& en & \textbf{69.2} & 71.9 \\
& es & 19.9 & \textbf{22.8} \\
& fr & 43.3 & \textbf{45.8} \\
& it & 43.6 & \textbf{46.6} \\
\hline
\multirow{10}{*}{XNLI}
& ar & 32.8 & \textbf{33.2} \\
& de & 50.5 & \textbf{51.9} \\
& en & \textbf{55.4} & 55.2 \\
& es & 50.3 & \textbf{52.3} \\
& fr & 51.1 & \textbf{52.3} \\
& hi & 43.9 & \textbf{46.5} \\
& ru & 43.8 & \textbf{47.3} \\
& tr & 44.8 & \textbf{45.6} \\
& vi & 45.1 & \textbf{45.3} \\
& zh & 33.0 & \textbf{35.5} \\
\hline
\multirow{7}{*}{PAWS-X}
& de & 63.3 & \textbf{64.7} \\
& en & 65.6 & \textbf{65.7} \\
& es & 58.4 & \textbf{61.1} \\
& fr & 58.5 & \textbf{58.6} \\
& ja & \textbf{56.1} & 53.2 \\
& ko & 49.3 & \textbf{51.2}\\
& zh & 50.8 & \textbf{53.1} \\
\bottomrule
\end{tabular}
\caption{Per-language accuracy (\%) for each downstream task using Wanda and M-Wanda on Llama-8B.}
\label{tab:per_language_accuracy}
\end{table*}

\end{document}